\documentclass[journal]{IEEEtran}
\usepackage{blindtext}
\usepackage{graphicx}
\usepackage{color}
\usepackage{gensymb}
\usepackage{amsmath}
\usepackage{amssymb}
\usepackage{algorithm}
\usepackage[noend]{algpseudocode}
\def\BState{\State\hskip-\ALG@thistlm}
\ifCLASSINFOpdf
\else
\fi
\begin{document}

\title{Minor Privacy Protection Through Real-time Video Processing at the Edge}

\author{
\IEEEauthorblockN{Meng Yuan${^a}$, Seyed Yahya Nikouei${^a}$, Alem Fitwi${^a}$, Yu Chen${^a}$, Yunxi Dong${^b}$}
\IEEEauthorblockA{\\${^a}$Dept. of Electrical and Computer Engineering,
Binghamton University, Binghamton, NY 13902, USA \\
${^b}$Martin J. Whitman School of Management, Syracuse University, Syracuse, NY 13244, USA \\
Emails: \{myuan3, snikoue1, afitwi1, ychen\}@binghamton.edu, ydong17@syr.edu}
}

\maketitle

\begin{abstract}

The collection of a lot of personal information about individuals, including the minor members of a family, by closed-circuit television (CCTV) cameras creates a lot of privacy concerns. Particularly, revealing children's identifications or activities may compromise their well-being. In this paper, we investigate lightweight solutions that are affordable to edge surveillance systems, which is made feasible and accurate to identify minors such that appropriate privacy-preserving measures can be applied accordingly. State of the art deep learning architectures are modified and re-purposed in a cascaded fashion to maximize the accuracy of our model. A pipeline extracts faces from the input frames and classifies each one to be of an adult or a child. Over 20,000 labeled sample points are used for classification. We explore the timing and resources needed for such a model to be used in the Edge-Fog architecture at the edge of the network, where we can achieve near real-time performance on the CPU. Quantitative experimental results show the superiority of our proposed model with an accuracy of 92.1\% in classification compared to some other face recognition based child detection approaches. 

\end{abstract}


\IEEEpeerreviewmaketitle

%%%%%%%%%%%%%%%%%%%%%%%%%%%%%%%%%%%%%%%%%%%%%%%%%%%%%%%%%%%%%%%%%%%%%%%%%%%%%%%%%%%%%%%%%%%%%%%%%%%%%%%%%%%%%%%%%%%%%%%%%%%%%%%%%%%%%%%%%%%%%%%%%%%%%%%%%%%%%%%%%%%%%%%%%%%%
%%% Start %%%

\section{Introduction}
\label{sec:intro}

With increasingly ubiquitous deployment of smart surveillance cameras throughout the cities where majority of population lives, privacy issues are getting into focus \cite{cavallaro2007privacy, dufaux2011video}. Privacy often defines the boundaries to limit access to an individual’s private information and body. Today, we live in an information society where vast quantities of data about us are gathered and analyzed through automated processes and cameras. A lot of private attributes and personal information about individuals are collected by closed-circuit television (CCTV) cameras and streamed to remote cloud servers and viewing stations with no privacy protection mechanism enforced \cite{taylor2010spy}. 

Initially, these surveillance cameras were deployed for public safety purposes and to provide concrete evidence for forensics analysis \cite{chen2017enabling, xu2018real}. The user may vary from the public safety authorities, law enforcement agents to a house owner. With the transmission of the unprotected video through the communication network the video may be subject to attacks. As a consequence, these large amount of data collected by the cameras could be intercepted and abused by adversaries. For example, a man in the middle who can view the raw frames is considered as a breach of the privacy \cite{hessler2006museum, kumar2015promoting, newton2005preserving}. This has caused the public to be more concerned and to ask for change in the way video surveillance works \cite{fitwi2019agent, lyon2010surveillance}.

Specifically, the practice of mass-surveillance can have a profound effect on the understanding of minors about privacy in their later lives \cite{waters2018effects}. Usually children learn through experience; hence, they should grow up in an environment where privacy is practiced if they are to learn what privacy is and how it works. Besides, many argue that the right experience of privacy is very important to a child’s future success and good decision-making in setting correct safely measures and social privacy boundaries. As a result, today’s pervasive surveillance systems must have means to protect children’s privacy. 

Privacy protection is one of the active research areas in the rise of Internet of Things (IoT) \cite{yang2017survey}, where a huge number of sensors and low powered processors are going to be connected to the network with none or minimal security measures. One of the more important aspects of this research is to protect the identity of the people in case the data is compromised. Any effort to address the privacy problems in a surveillance system must have techniques for identifying private attributes on images and for protecting these attributes \cite{fitwi2019no, nikouei2020ivise}. 

Private attributes like face are detected through the use of machine learning or deep learning networks \cite{fitwi2019no}. Following detection, these private attributes of individuals are scrambled using apropos cryptographic schemes. These schemes ensure that video streams are not accessed by means of interception and abused by unauthorized people while being transmitted from the cameras to the cloud servers and viewing centers. Among variant privacy preserving requirements, minor children's identity and face protection is essential to every family to protect the minors from attackers or abusers \cite{berson2006children, lwin2008protecting, shmueli2010privacy}. 

In this work, we propose a novel Minor Privacy protection solution using Real-time video processing at the Edge (MiPRE). In MiPRE, the video is checked by the smart cameras with Deep Neural Network (DNN) to detect children's faces, and then a lightweight blurring algorithm is called to scramble the faces before the raw video is transmitted through network to the consumer or the storage drive. Therefore, the MiPRE scheme protects the privacy of minor children by securely denaturing their faces. In this paper, we present the face detection and recognition model of the MiPRE system, which categorizes the tested images into adults and children. More specifically, the face detection method employs the Multitask Convolutional Neutral Network (MTCNN) \cite{zhang2016joint} to detect and align the faces. The face recognition is realized using the FaceNet model \cite{schroff2015facenet}, which is designed by Google group, was employed to extract the feature embedding of children’s faces.

The rest of this paper is organized as follows. In section \ref{background} we discuss several methods for children detection as well as the historical efforts in the detection and recognition of human faces. Section \ref{sec:arch} presents the system architecture of our MiPRE scheme and its function blocks are discussed in detail, including the multi-step pipeline face detection and children recognition. Section \ref{sec:experimental} reports the model training process and the performance of the children detection. Finally, Section \ref{sec:conclusions} concludes this paper.

%%%%%%%%%%%%%%%%%%%%%%%%%%%%%%%%%%%%%%%%%%%%%%%%%%%%%%%%%%%%%%%%%%%%%%%%%%%%%%%%%%%%%%%%%%%%%%%%%%%%%%%%%%%%%%%%%%%%%%%%%%%%%%%%%%%%%%%%%%%%%%%%%%%%%%%%%%%%%%%%%%%%%%%%%%%%%%%%%%%%%%%%%%%%%%%%%%%%%%%%%%%%%%%%%%%%%%%%%%%%%%%%%%%%%%%%%%%%%%%%%
\section{Related Work}
\label{background}

%\subsection{Face Recognition based Age Recognition}

With the development of machine learning, computers are becoming more widely used, which reduces manual workload and guarantees high recognition rate \cite{amos2016openface, cox2011beyond}. In recent years the community also witnesses the migration of powerful machine learning algorithms to the IoT environments by developing lightweight solutions \cite{nikouei2018lcnn, nikouei2018smart, wu2017container, xu2018real}.

In the field of face recognition, research is mainly focused on two aspects, namely authentication \cite{fathy2015face, vazquez2016face} and recognition \cite{sharif2016accessorize, yaman2018comparison}. In the face recognition process, whether it is recognition or authentication, a well-known method is top-bottom approach where face rectangle is first detected, features form the face are extracted and finally a comparison is made \cite{uiboupin2016facial}. In face recognition, human age recognition is a well-studied issue \cite{li2018distance}. Classifiers are trained to detect the age of the subject or to predict the facial appearance in certain age group. Building on the state of the art architectures, we present a unique decentralized method for children detection which performs the most accurate.

Face recognition-based age recognition is the process of extracting age-related facial features, create an age classification model \cite{bhattacharya2019survey, zhou2018age}. Then, use this model to evaluate the age range of given person to categorise this person into different age groups. The ability to build a model through face recognition is because human aging and changes are not changed by human willpower. This is a complex process that is related to the health and status of people's living environment, etc. %Hence, we designed a children’s face recognition model based on the face features.

Although the research on face recognition started earlier, there are few studies on the establishment of children classification models. Today’s top-performing techniques of face recognition are based on Multi-task convolutional neural networks. Both Facebook’s DeepFace \cite{taigman2014deepface} and Google’s FaceNet \cite{schroff2015facenet} architectures have the highest accuracy. DeepFace uses 6 conv. DeepFace uses 6 conv. layers followed by two FC layers that are used to detect and map a face in 3-D space and to map 67 fiducial points on the face. Facenet approach is to detect faces that belong to the same person using illumination and Pose invariance architecture. MTCNN and FaceNet are employed in our model to reach a better results compared with state of the art techniques.

%%%%%%%%%%%%%%%%%%%%%%%%%%%%%%%%%%%%%%%%%%%%%%%%%%%%%%%%%%%%%%%%%%%%%%%%%%%%%%%%%%%%%%%%%%%%%%%%%%%%%%%%%%%%%%%%%%%%%%%%%%%%%%%%%%%%%%%%%%%%%%%%%%%%%%%%

\section{MiPRE: Minor Privacy Protection at the Edge}
\label{sec:arch}

%\subsection{System Overview}

\begin{figure}[t]
    \centering
        \includegraphics[width=0.49\textwidth]{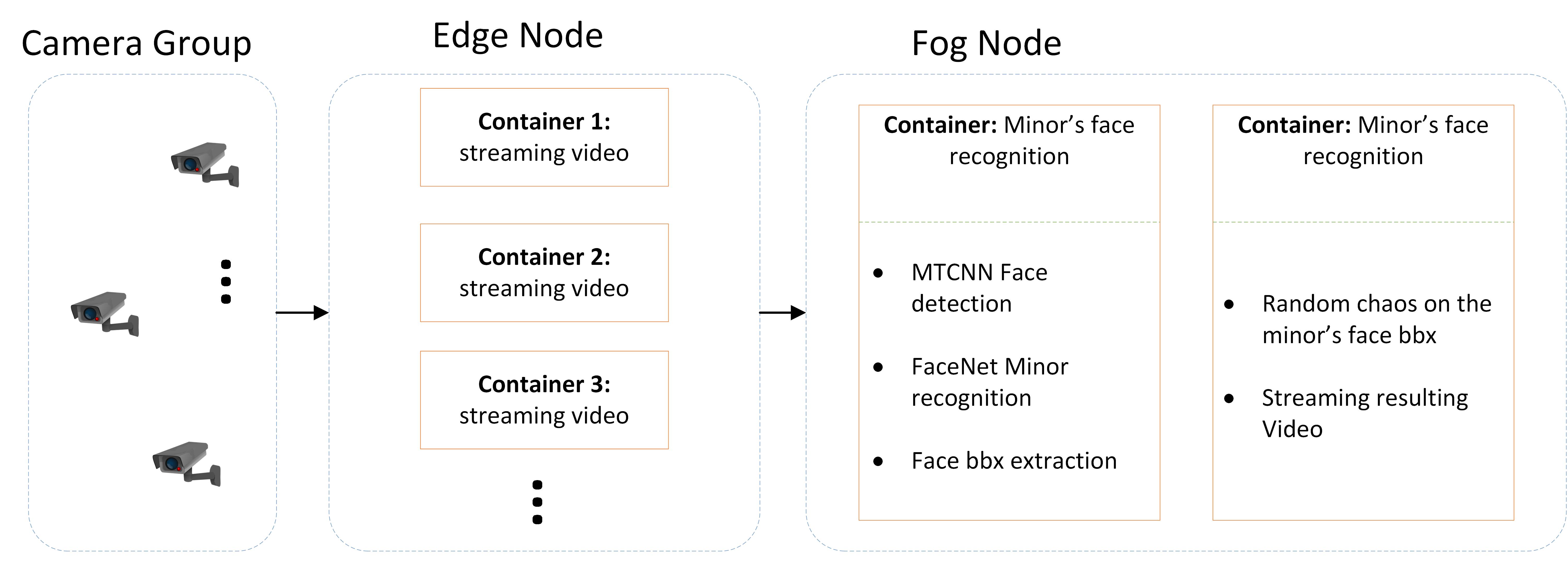}
    \caption{MiPRE system architecture.}
    \label{fig:archi}
    \vspace{-10pt}
\end{figure}

Figure \ref{fig:archi} presents the architecture of our MiPRE system. It consists of three major function blocks: (1) face detection using a multi-step pipeline model, (2) face recognition based on the extracted features and separate faces of children from adults, and (3) face scrambling to protect children's privacy. Each module is implemented in a docker container which promises scalability and faster updates in parts of the system using microservices architecture \cite{nagothu2018microservice, nikouei2019decentralized}. The design rationales and technical details of face detection and face recognition are presented in the following sections. The face scrambling for privacy protection is beyond the scope of this paper, interested readers may find the complete description of our MiPRE scheme in \cite{fitwi2020minor}.

\subsection{Face Detection}
\label{sec:model}

While there are many face detection methods, such as Dilb or OpenFace face detection, MTCNN (Multitask Convolutional Neutral Networks) is adopted in this work for two reasons. On one hand, it achieves a high detection accuracy, and on the other hand, FaceNet model has already provided MTCNN interface to detect faces.

Basically, MTCNN is a deep learning model for face detection based on a multi-task cascaded Convolutional Neural Network (CNN). It exploits the inherent correlation between detection and alignment to boost up its performance. In particular, to predict face and landmark locations in a coarse-to-fine manner, the framework used in this paper leverages a cascaded architecture with three stages of carefully designed deep conv. networks \cite{taigman2014deepface, zhang2016joint}. 

\begin{figure}[t]
    \centering
        \includegraphics[width=0.48\textwidth]{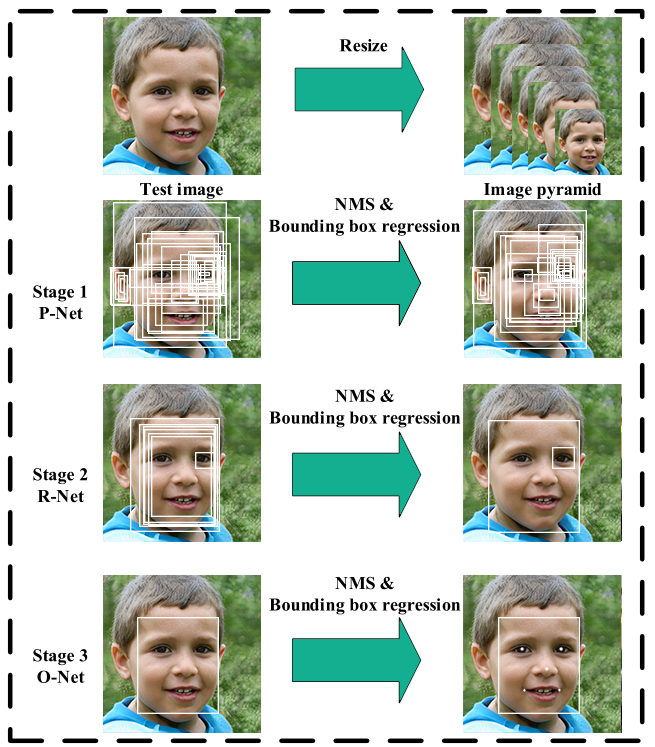}
    \caption{MTCNN: steps the cascaded architecture takes to ensure best performance in human face detection and bounding box regression.}
    \label{fig:MTCNN}
    \vspace{-10pt}
\end{figure}

Given an input image, an image pyramid is made by re-scaling the image into different scales through a bi-linear interpolation. This step insures scale invariation. Figure \ref{fig:MTCNN} shows an example, in the MTCNN the three cascaded stages follow scaling step: 

\begin{itemize}
    \item \emph{P-Net}: It is a full convolutional neural network (FCN). The feature map obtained by forward propagation is a 32-dimensional feature vector at each position. It is used to determine whether or not grid cells of $12\times12$ contain a face. If a grid cell contains human face, the Bounding Box of the human face is regressed, and the Bounding Box corresponding to the area in the original image is further obtained. The Bounding Box with the highest score is retained by a Non-maximum suppression (NMS) step and all of the other Bounding Boxes with an excessively large overlapping area are removed.
    
    \item \emph{R-Net}: It is a simple CNN stage. Similar to the last stage (O-Net), the $24\times24$ and the resulting Bounding Box area is up-scaled to $48\times48$. It is then given to the R-Net stage to have the highest detection confidence of Bounding Box detection and facial landmark extraction.
    
    \item \emph{O-Net}: O-Net is for higher accuracy. It is a simple CNN, the Bounding Box that P-Net step produces may or may not contain a human face. This box as well as the $12\times12$ input is first up-scaled using a bilinearly interpolation method to $24\times$24, which is then used as the input to the O-Net to determine whether a human face exists. If a human face is contained, the Bounding Box is regressed, which is also followed by the NMS step.
\end{itemize}

\begin{figure}[t]
    \centering
        \includegraphics[width=0.48\textwidth]{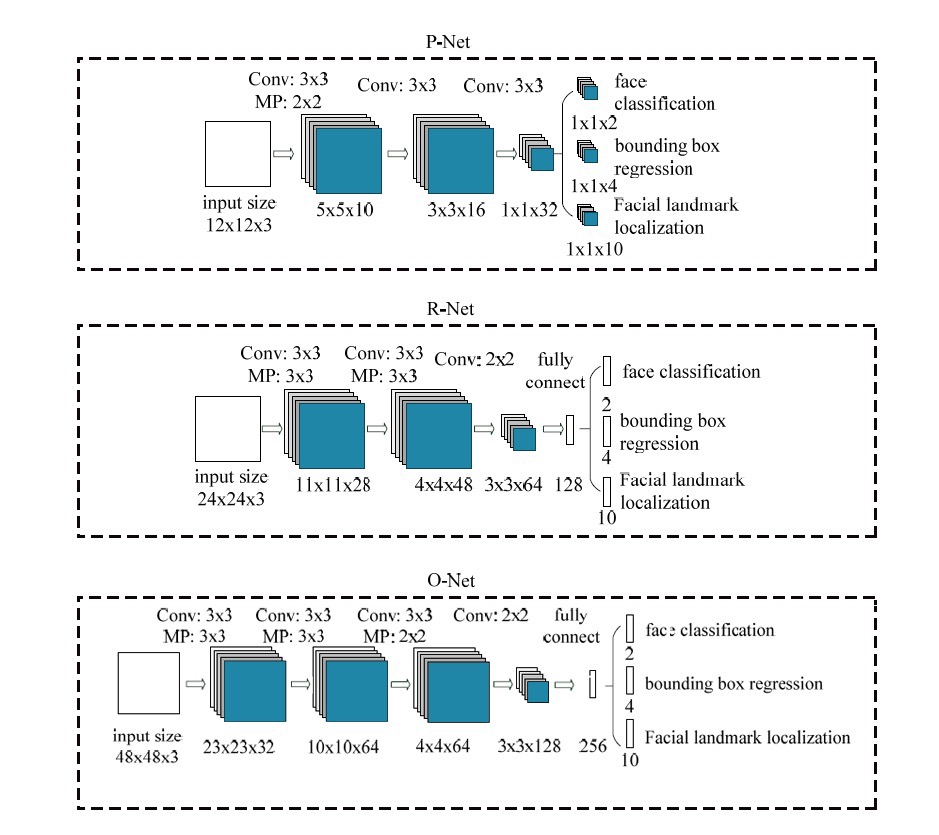}
    \caption{MTCNN: Stage architecture of the model used for face detection and landmark extraction.}
    \label{fig:arch}
    \vspace{-10pt}
\end{figure}

Figure \ref{fig:arch} presents the architecture of the layers used in each of the stages in the cascaded MTCNN model. Each step uses different sizes of Conv. filters and different number of layers to produce the same class of results. The outputs are in three categories. The face classification score is presented as the first set of outputs using two neurons. One for the presence of a face and the other as the score. Another part of the output is the bounding box regression where four neurons present the upper left and lower right of the bounding box as $dx_1, dy_1, dx_2, dy_2$. Facial landmark localization regresses the position of five points of left eye, right eye, nose, left mouth corner, and right mouth corner.

During the training phase, the three networks will use the landmark positions as supervised signals to guide the learning of the network. In the prediction phase, however, P-Net and R-Net only conduct face detection and do not output landmark positions because they are inaccurate in these pases. The landmark position is only obtained the O-Net. Bounding box and landmarks coordination outputs are normalized relative to the input image.

%\noindent \textbf{Training:} 
As mentioned above, there are three tasks that MTCNN archives. Namely face classification, bounding box regression and facial landmark localization. Thus, the loss function of the algorithm also has three sections. Due to the limited space, here is a highlight of the key points, readers interested for more details are referred to \cite{taigman2014deepface}.

Cross-entropy loss function is employed for face classification as shown in Eq. (\ref{eq:cross}):

\begin{equation}
 L^{det}_i = -(y^{det}_ilog(p_i)+(1-y^{det}_i)(1-log(p_i)))
 \label{eq:cross}
\end{equation}

\noindent where the $y^{det}_i$ shows the ground truth for object $i_{th}$ and the $p_i$ is the network output for the face detection. 

Next is the bounding box regression loss where the euclidean distance loss function is employed as seen in Eq. (\ref{eq:bbx}):

\begin{equation}
 L^{bbx}_i = ||(\hat{y}^{bbx}_i - y^{bbx}_i||^2_2
 \label{eq:bbx}
\end{equation}

Lastly, the same regression loss is used for each of the $l$ landmark for each samples $i$ as mentioned in Eq. (\ref{eq:land}):

\begin{equation}
 L^{landmark}_i = ||(\hat{y}^{landmark}_i - y^{landmark}_i||^2_2
 \label{eq:land}
\end{equation}

%%%%%%%%%%%%%%%%%%%%%%%%%%%%%%%%%%%%%%%%%%%%%%%%%%%%%%%%%%%%%%%%%%%%%%%%%%%%%%%%%%%%%%%%%%%%%%%%%%%%%%%%%%%%%%%%%%%%%%%%%%%%%%%%%%%%%%%%%%%%%%%%%%%%%%%%%%%%%
\subsection{Children Faces Recognition}
\label{sec:platfrom}

There are several ways to compare the similarity of two images. The euclidean distance metric is one of the most used one, because of the ease in implementation and no expensive computation. Given a feature map where the features are extracted from the face, this metric is going to show the similarity in the features between the feature set and a known set. This idea is the back bone of this section where we are going to feed the faces that are extracted in the face detection step to the FaceNet and compare the resulting feature map with datasets that are know positive and negative images of children's faces. A similarity threshold is then picked to give a final label to the face. 

FaceNet is a universal system that can be used for face authentication, recognition and clustering. FaceNet's approach is to learn to map images to an Euclidean space through CNNs. Spatial distance is directly related to the similarity of pictures. Different images of the same person have a small spatial distance, and images of different people have a larger distance in space. As long as the mapping is determined, the related face recognition task will be simple \cite{jose2019face}.

\begin{figure}[t]
    \centering
        \includegraphics[width=0.48\textwidth]{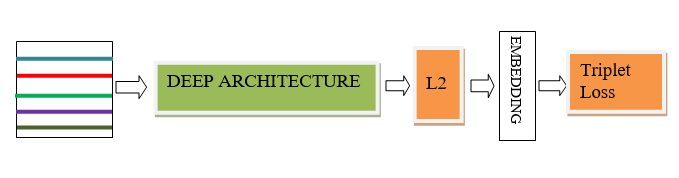}
    \caption{Model structure: This network consists of a batch input and output layer and a deep CNN followed by L2 normalization, which results in the face embedding. This is followed by the triplet loss during training.}
    \label{fig:model}
    \vspace{-10pt}
\end{figure}

Currently, existing DNN-based face recognition models use a FC classification layer. The middle layer in the FC layers, after the Conv. layers, or the last Conv. layer is a vector map of the face image. The FC classifier layer is then placed on top of this vector map. The disadvantages of such methods are indirectness and inefficiency. In contrast, FaceNet directly uses the loss function of triplets-based Large Margin Nearest Neighbor (LMNN) to train the neural network, and the network directly outputs a 128-dimensional vector space. The triplets we selected contain two matching face thumbnails and one non-matching face thumbnail. The goal of the loss function is to distinguish positive and negative classes by distance boundaries as shown in Fig. \ref{fig:model}

\begin{figure}[t]
%    \vspace{-10pt}
    \centering
        \includegraphics[width=0.48\textwidth]{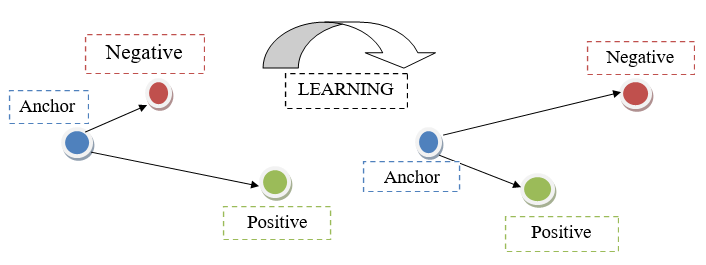}
    \caption{Training goal of the FaceNet network.}
    \label{fig:train}
    \vspace{-10pt}
\end{figure}

The purpose of the model is to embed the 2-D face image $X$ into the Euclidean space with $D$ dimensions where $f(X) \in R^d$. In this vector space the anchor image of a face $x^a_i(anchor)$ is close to other images with the same facial expressions ($x^p_i(positive)$) and far from faces with different characteristic ($x^n_i(negative)$). As illustrated by Fig. \ref{fig:train}, the training process migrates the network's behavior pattern from the left side to the right side. 

To reach this goal a triplets loss function is calculated from triplet of three pictures. The triplet is composed of Anchor (A), Negative (N), and Positive (P) images. Any image can be used as a base point (A), then images which have the same facial characteristics are its (P) and images that do not share the same characters are considered as its (N). Triplets Loss minimizes the distance between an anchor and a positive, both of which have the same identity, and maximizes the distance between the anchor and a negative image. Thus, the loss function can be formulated as:

\begin{equation}
 L = \sum^N_i [||f(x^a_i) - f(x^p_i)||^2_2 - ||f(x^a_i) - f(x^n_i)||^2_2 + \epsilon]
 \label{eq:loss}
\end{equation}

\noindent where $\epsilon$ is the safe boundary between the positive and negative. 

Theoretically speaking, the best images for training purposes are the ones with the highest distance between the (A) and (N) and lowest between the (A) and (P). However, in practice this approach creates local minimum and global solution is not going to be reached. A remedy to this problem is to select all positive image pairs in a mini-batch, which can make the training process more stable. For the selection of the (N), on the other hand, as long as the Eq. (\ref{eq:nonequal}) is met the network is going to be trained. 

\begin{equation}
 ||f(x^a_i) - f(x^p_i)||^2_2 < ||f(x^a_i) - f(x^n_i)||^2_2
 \label{eq:nonequal}
\end{equation}

\begin{figure*}[t]
    \centering
        \includegraphics[width=0.98\textwidth]{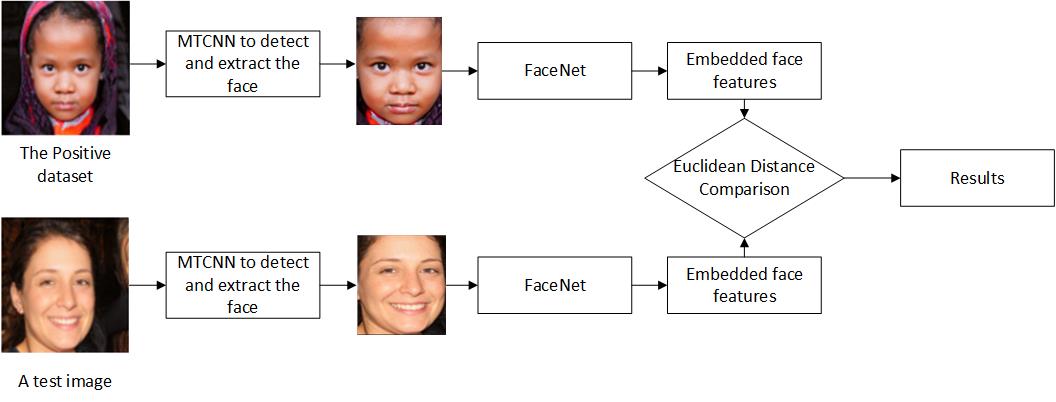}
    \caption{Dataflow in our model shows how we use MTCNN to detect and crop the faces from each input image and then use FaceNet to calculate the euclidean distance between the anchor face and positive and negative images to detect children's faces.}
    \label{fig:diagram}
    \vspace{-10pt}
\end{figure*}

\noindent \textbf{Training:} in order to train this network we used 10,000 images of children and 10,000 images of adult faces. The children ages are variable between 6 to 14 years old. For positive selection we selected 100 children and 100 images of adult faces. The images are fed through the MTCNN to have a Bounding Box around the face. The face rectangle is then fed to the FaceNet for euclidean distance calculation. Figure \ref{fig:diagram} presents the data flow of our model for minor's face detection. 

In Fig. \ref{fig:diagram} the positive and negative dataset with which a test image is compared against is prepared before hand. This feature set is called Embedding dataset and has all of the feature maps from all 200 aforementioned images for comparison. Total inference time, thus, is divided to two parts: face detection time and feature comparison using the FaceNet network. More details will be presented in Section \ref{sec:experimental}.

%%%%%%%%%%%%%%%%%%%%%%%%%%%%%%%%%%%%%%%%%%%%%%%%%%%%%%%%%%%%%%%%%%%%%%%%%%%%%%%%%%%%%%%%%%%%%%%%%%%%%%%%%%%%%%%%%%%%%%%%%%%%
\section{Experimental Results}
\label{sec:experimental}

\subsection{Experimental Setup}

The multi-level MiPRE architecture is tested on a x86 based CPU. The model is to be executed on edge server grade hardware that are more powerful in nature than low-powered edge devices. In this context we consider a laptop or PC to be edge server grade and devices such as raspberry PI to be edge modules. Specifically, the MiPRE model is tested on a AMD Ryzen 7 2700X processor with 3.7GHz based clock with 8 cores. The system has dual 8GB memory modules and is running a windows 10. During inference, we observed average of 18\% CPU utilization which is acceptable considering the need to connect several edge nodes to an edge server. On the other hand, memory utilisation of the process is higher at about 10GB on average. Although there is no surprises in memory usage because of loading several stages of CNN models, it should be considered when deploying the model. 

%\subsection{Performance}

\subsection{Accuracy of Face Recognition}

We compared the accuracy of our MiPRE model to the state of the art models for facial components based on the age. Table \ref{accu} shows the comparison. The approach reported in \cite{otto2012does} tries to divide faces to multiple components and use their changes as the features to indicate the age of the subject face. Levi and his colleagues \cite{levi2015age} use a CNN to classify primary objects in an image between gender and age. Meanwhile, a rule based method has been also proposed that divides the image into sections and implement privacy measures based on rule sets \cite{teixeira2014rule}. The last method we compared with is \cite{du2011face} which tries to extract facial features and accurately detect the age of each. As shown by Table 1, our MiPRE scheme achieves a better performance in terms of accuracy than of these reported efforts.

\begin{table}
\begin{center}
\begin{tabular}{|c|c|}
\hline
Model & Accuracy \\
\hline\hline
Otto et al. \cite{otto2012does} &  81.27 \% \\
\hline
Levi et al. \cite{levi2015age} & 84.7 \% \\
\hline
Teixeira et al. \cite{teixeira2014rule} & 91.14 \% \\
\hline
Du et al. \cite{du2011face} & 79.24 \% \\
\hline
MiPRE & 92.1 \% \\
\hline
\end{tabular}
\end{center}
\caption{Accuracy of face classification based on the age. Our multi staged model has achieved a higher average accuracy.}
\label{accu}
\end{table}

Table \ref{nums} shows the ratio at which the multi staged model we proposed in MiPRE works. The model achieves a miss detection rate of 7.9\% in the testing set, as 158 out of 2000 images were mis-categorized. Meanwhile, the detection rate of 92.1\% is among the best.

\begin{table}
\begin{center}
\begin{tabular}{|c|c|c|c|}
\hline
Miss detection & Miss detection rate & True Detection & Detection rate\\
\hline\hline
 158 & 0.079 & 1842 & 0.921 \\
\hline
\end{tabular}
\end{center}
\caption{Miss classification rate based on the 2000 images that are used for testing.}
\label{nums}
\end{table}

\begin{figure}[t]
    \centering
        \includegraphics[width=0.45\textwidth]{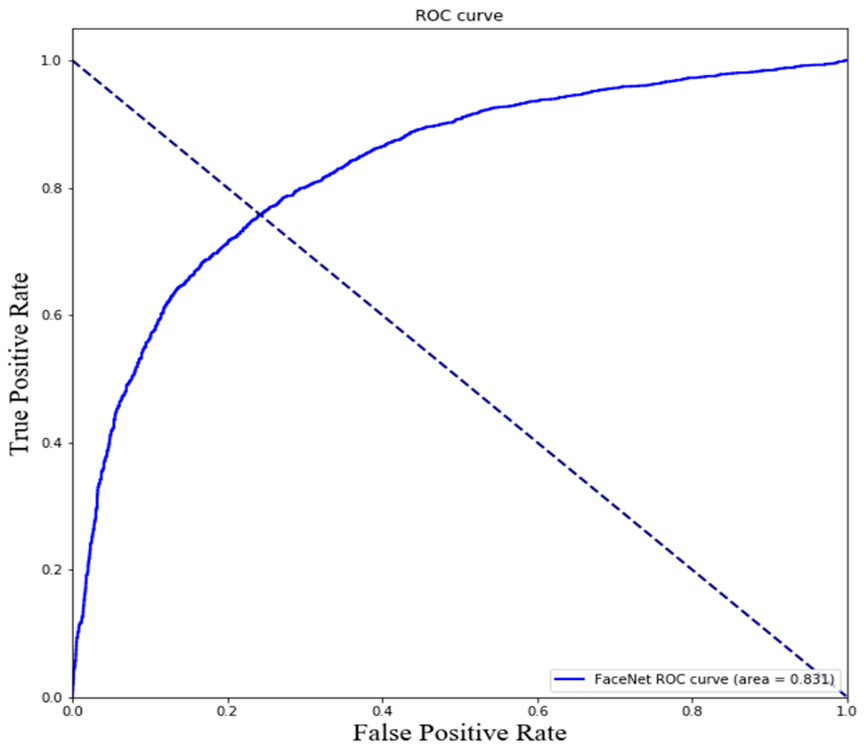}
    \caption{ROC curve.}
    \label{fig:roc}
    \vspace{-10pt}
\end{figure}

Figure \ref{fig:roc} is the ROC curve that shows detection rate, shown as True Positive Rate versus the False Positive rate. This curve gives some intuitive insight to the best possible threshold to be set for the child detection. A bigger area under this curb means that the system performs better with higher true positive and lower false positive rate. During implementation, for example, if the true positive rate of 0.7 is needed, then a 0.2 false positive rate is going to be expected. 

%%%%%%%%%%%%%%%%%%%%%%%%%%%%%%%%%%%%%%%%%%%%%%%%%%%%%%%%%%%%%%%%%%%%%%%%%%%%%%%%%%%%%%%%%%%%%%%%%%%%%%%%%%%%%%%%%%%%%%%%%%%%%%%%%%%
\section{Conclusions}
\label{sec:conclusions}

As the number of surveillance cameras increases, families are more concerned about the privacy of their members and their personal data. Child protection is a vital role of the parents and it is important to minimize unauthorized video appearance of the minor children. Particularly, face is one of the most powerful human identifying attributes and scrambling it can effectively anonymize individuals. In this work, a novel lightweight minor privacy protection scheme named MiPRE is proposed. Leveraging a multi-stage DNN based face recognition approach to detect children in the video and a lighweight chaos based face scrambling algorithm, the MiPRE scheme ensures de-identification of the minors at the edge of the network, before the video is streamed to the Internet. The MiPRE scheme is tested on a platform consisting of a smart camera and an edge server, the experimental results verified that the MiPRE scheme meets the design goal. It achieved a high accuracy in children face recognition, $92.1\%$. %, and is able to finish the face scrambling operation in 45 ms.  

Our on-going effort mainly focus on identifying other private attributes that have significant impacts on children privacy-preserving, which allows us to extend the coverage of the MiPRE scheme. Corresponding to additional computing capacities raised by features other than faces, we will continue investigating lightweight machine learning algorithms to fit the next version of the MiPRE scheme in the edge environments.

%%%%%%%%%%%%%%%%%%%%%%%%%%%%%%%%%%%%%%%%%%%%%%%%%%%%%%%%%%%%%%%%%%%%%%%%%%%%%%%%%%%%%%%%%%%%%%%%%%%%%%%%%%%%%%%%%%%%%%%%%%%%%%%%%%%%%%%%%%%%%%%%%%%%%%%%%%%%%%%%%%%%%%%%%%%%%%%%%%%%%%%%%%%%%%%%%%%%%%%%%%%%%%%%%%%%%%%%%%%%%%%%%%%%%%%%%%%%%%%%%%%%%%%%%%%%%%%%%%%%%%%%%%%%%%%%%%%%%%%%%%%%%%%%%%%%%%%%%%%%%%%%%%%%%%%%%%%%%%
%%% END %%%
\ifCLASSOPTIONcaptionsoff
  \newpage
\fi
\bibliographystyle{IEEEtranS}
\bibliography{L-CNN.bib}
\end{document}